\documentclass[10pt]{article}

\usepackage[english]{babel}
\usepackage{csquotes}
\usepackage{longtable}
\usepackage{ragged2e}
\usepackage{multirow}
\usepackage[hidelinks]{hyperref}
\usepackage{amsfonts}

\usepackage[margin=1in]{geometry}

\usepackage{graphicx}
\usepackage{float}      % enables [H]
\usepackage{placeins}   % enables \FloatBarrier
\usepackage{tabularx}
\usepackage{booktabs}   % nicer tables (optional)
\usepackage{caption}    % optional but helpful

\graphicspath{{figures/}{figures/}}

\usepackage{titlesec}
\titleformat{\section}
  {\normalfont\large\bfseries}
  {\thesection}{0.75em}{}

\titleformat{\subsection}
  {\normalfont\normalsize\bfseries}
  {\thesubsection}{0.75em}{}

\title{Patient foundation model for risk stratification in low-risk overweight patients}

\author{
Zachary N. Flamholz\textsuperscript{1*\textdagger}
\and
Dillon Tracy\textsuperscript{1*}
\and
Ripple Khera\textsuperscript{1}
\and
Jordan Wolinsky\textsuperscript{1}
\and
Nicholas Lee\textsuperscript{1}
\and
Nathaniel Tann\textsuperscript{1}
\and
Xiao Yin Zhu\textsuperscript{1}
\and
Harry Phillips\textsuperscript{1}
\and
Jeffrey Sherman\textsuperscript{1\textdagger}
}

\begin{document}
\maketitle % Output the title and abstract box
\noindent\textsuperscript{1}Zephyr AI, Inc., McLean, VA, USA\\
\textsuperscript{*}These authors contributed equally.\\
\textsuperscript{\textdagger}Co-corresponding authors: zach@zephyrai.bio, jeff@zephyrai.bio

%----------------------------------------------------------------------------------------
%	ARTICLE CONTENTS
%----------------------------------------------------------------------------------------
\section*{Abstract}
Accurate risk stratification in patients with overweight or obesity is critical for guiding preventive care and allocating high-cost therapies such as GLP-1 receptor agonists. We present PatientTPP, a neural temporal point process (TPP) model trained on over 500,000 real-world clinical trajectories to learn patient representations from sequences of diagnoses, labs, and medications. We extend existing TPP modeling approaches to include static and numeric features and incorporate clinical knowledge for event encoding. PatientTPP representations support downstream prediction tasks, including classification of obesity-associated outcomes in low-risk individuals, even for events not explicitly modeled during training. In health economic evaluation, PatientTPP outperformed body mass index in stratifying patients by future cardiovascular-related healthcare costs, identifying higher-risk patients more efficiently. By modeling both the type and timing of clinical events, PatientTPP offers an interpretable, general-purpose foundation for patient risk modeling with direct applications to obesity-related care and cost targeting.

\section*{Introduction}
Overweight and obesity affect more than 70\% of U.S. adults \cite{fryar2020prevalence} and are associated with increased risk for a wide range of chronic conditions, including cardiovascular disease (CVD) \cite{obesityCVD2024}, type 2 diabetes \cite{diabetesObesity}, osteoarthritis \cite{murphy2008lifetime}, and certain cancers \cite{Recalde2023}. While the health \cite{NEJMobesity2015} and economic \cite{dall2024assessing} burdens of obesity are well established, treatment strategies remain imprecise, particularly for individuals classified as low-risk by traditional clinical metrics. The advent of glucagon-like peptide-1 receptor agonists (GLP-1s), such as semaglutide and tirzepatide, has shown promise in promoting weight loss and reducing cardiovascular events in high-risk patients \cite{SELECT2023, SUMMIT2025}. However, in lower-risk populations, the largest patient group eligible for GLP-1s \cite{semaglutideShi}, the number needed to treat (NNT) to prevent a single major adverse cardiovascular event remains high, raising concerns about the cost-effectiveness and broad deployment of these medications \cite{KOMATSOULIS2025100587}. To expand access to GLP-1 medications, more refined risk stratification tools are needed to identify which low-risk individuals are likely to benefit most from early pharmacologic intervention.\\
\newline
Traditional risk prediction models used in primary CVD prevention strategies, such as ASCVD Risk Calculator \cite{ASCVDriskCalcualtor}, QRISK3 \cite{QRISK3}, and PREVENT \cite{PREVENT2023}, focus on individual outcomes such as myocardial infarction, stroke, heart failure or some combination thereof. While such models are clinically interpretable, they do not account for the interdependence and temporal ordering of multiple obesity-associated conditions, which often co-occur and compound over time \cite{wang2014AMIAjointrisk}. As a result, they fail to capture the joint distribution of risk across a patient’s full trajectory. This is particularly limiting in evaluating interventions like GLP-1s, whose benefits may be distributed across a range of endpoints including metabolic \cite{sema2021weight, tirz2022weight}, cardiovascular \cite{SELECT2023, SUMMIT2025, KOMATSOULIS2025100587}, musculoskeletal \cite{glp12024arthritis}, and behavioral health \cite{glp12025AUD} outcomes. A more holistic representation of patient risk that reflects both the diversity and timing of potential future events is needed to inform precision prevention strategies and improve patient selection for emerging therapies.\\
\newline
Recent advances in artificial intelligence have demonstrated the value of foundation models trained on large-scale patient data to learn general-purpose representations that support a wide range of downstream clinical tasks. Often referred to as Foundation Models for Electronic Medical Records (FEMRs), these models aim to encode a patient’s longitudinal history into representations that capture patterns of illness, care, and disease progression over time. Existing FEMRs include both encoder-only architectures (e.g., CLMBR \cite{CLMBR2021Steinberg}, Med-BERT \cite{rasmy2021medbert}) and encoder–decoder frameworks (e.g., ETHOS \cite{renc2024ethos}, TEE4EHR \cite{karami2024tee4ehr}, TransformEHR \cite{transformehr2023}), and typically treat patient histories as sequences of discrete events. While many of these models incorporate some representation of time, such as age, time since last visit, or visit order, they do not explicitly model the distribution of event timing. Instead, time is treated as a static or contextual input feature. This limits their ability to fully capture the irregular and asynchronous nature of real-world clinical trajectories.\\ 
\newline
Temporal point processes (TPPs) offer a natural and flexible framework for modeling such trajectories \cite{yan2019tpp_survey}. A TPP is a modeling framework for sequences of events (``marks'') drawn from a discrete set of types (Figure 1A). It defines a conditional intensity function, which computes the probability per unit time of event occurrences given a certain history, or context. To learn this function from data, we take the first event in a sequence and try to predict its arrival time by sampling from some parametric distribution of interevent times (log-normal, for example) for some choice of parameters.  We append the predicted event to the context. This operation is folded over the training sequence and a loss is computed on the predicted sequence compared to the ground truth sequence.  Note that the model is modeling a joint conditional probability distribution, not outcomes per se, which are the realization of a sampling procedure on that distribution. TPPs therefore explicitly model the distribution of interevent times, enabling more accurate representation of evolving clinical risk and are well-suited to patient data that unfolds at irregular intervals where the timing and order of events carry important meaning (Figure 1B). Unlike traditional models that predict one outcome at a time, TPPs learn the overall structure of a patient's clinical history allowing them to generate a comprehensive risk profile over an aggregate of many different events. Event predictions that are out-of-distribution with respect to a given patient's history, representing new diagnoses or treatment starts, are relatively common. Applications of TPPs to patient data have demonstrated their generative \cite{enguehard2020neural, bhave2021popcorn} and predictive \cite{TPPicu} capacities but have not focused on patient representation learning.\\
% figure 1
\begin{figure*}[!htb]
\centering
\includegraphics[width=1.0\textwidth]{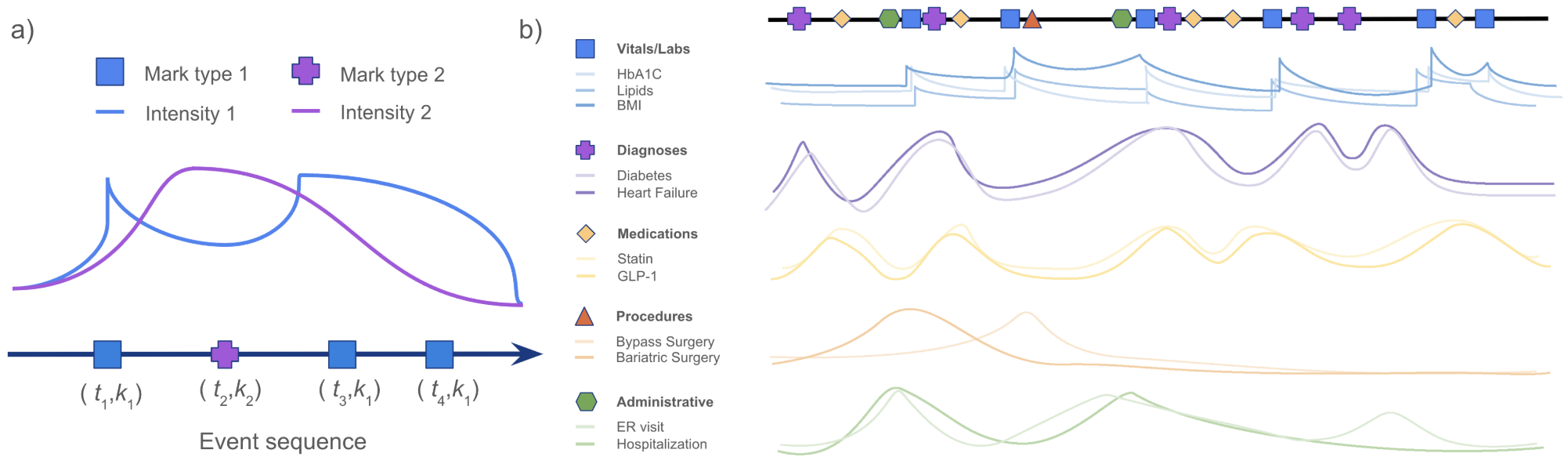}
\caption{\textbf{A.} A temporal point process is an event sequence defined over some interval $[0, T)$.  Events arrive at real-valued timestamps on this interval.  Events may 
have classes associated with them (also known as ``marks'') but not magnitudes (at least not out-of-the-box). Events have no duration. The process is presumed to be autoregressive, that is, past performance is some indication of future behavior. \textbf{B.} A TPP model learns a conditional intensity function, which represents the joint probability distribution over all events. Interdependencies between all event types are potentially captured. There are no restrictions on data spacing    or sparsity.}
\label{fig:tpp-schematic}
\end{figure*}
\newline
In this study, our objective was threefold. First, we evaluated TPP for patient representation learning from observational and claims data. We enhance out-of-the-box neuronal TPP for the purpose of FEMR modeling by utilizing prior knowledge in the form of pre-trained clinical term representations in TPP training, incorporating patient demographic data as static contextual information for event prediction, and utilizing the numeric information associated with some clinical events, for example lab test values. Second, we assessed the generalizability of a TPP representations to predict a range of clinically and economically meaningful endpoints related to overweight/obesity. Finally, we demonstrated that the model's ability to anticipate obesity-linked health events allows for risk stratification at the population level in the form of predicted future costs. Together, these findings validate our PatientTPP framework not only as a predictive engine, but also as a clinically grounded tool for healthcare decision making.

\section*{Methods}
\subsection*{Data source}
All patient histories were sourced from Optum’s de-identified Market Clarity Data (Optum\textsuperscript{\textregistered} Market Clarity), an integrated dataset that includes medical and pharmacy claims, along with electronic health records. It links clinical data such as lab results, vital signs, diagnoses, procedures, and data derived from unstructured clinical notes via natural language processing with historical administrative claims that encompass pharmacy, physician, facility, and medication information. The dataset is de-identified according to the HIPAA Privacy Rule's Expert Determination method and managed under Optum\textsuperscript{\textregistered} data use agreements \cite{hhs_deidentification_2012}. 
\subsection*{Data processing}
Optum\textsuperscript{\textregistered} Market Clarity data tables were converted to patient sequences by the following procedure: (1) all diagnoses, prescriptions, and labs were collected from claims and observations with time stamps, (2) codes were mapped to aggregated concept names, and (3) time was binned into quarter year intervals. The final patient data table for modeling was a fact-style table consisting of columns: patient identifier, interval, concept, and value. Value is a field that can be associated with some concepts, such as lab test results.
\subsection*{Temporal point process model}
PatientTPP is trained on clinical trajectories or ``sequences'' constructed from electronic health records and claims data. The analysis under discussion covers over 500,000 patient histories; patients were observed for the entirety of their histories prior to January 1, 2024. Sequences in this corpus begin no earlier than 2006.\\
\newline
Patients included for training were overweight or obese prior to January 1, 2021 but were excluded if they had diabetes, end stage renal disease, or cancer anytime before that date. Given our interest in modeling obesity-associated risk, we limited event types to a set of relevant conditions, lab tests, and treatments. In total, we include 3 demographic variables, 45 ICD-10-CM diagnoses, 8 LOINC observations, and 9 NDC medications (Supp. Data 1). \\
\newline
A neural temporal point process model implementation consists of three primary components: an event encoder, a history encoder, and an event decoder. The event encoder maps each timestamped clinical event into a fixed-length embedding. The history encoder summarizes the sequence of prior events to provide temporal context. While it is possible to use conventional recurrent architectures such as RNNs or LSTMs, we found the best results using a multi-headed attention-based framework derived from the Attentive Neural Hawkes Process (AttNHP) \cite{yang2022transformerembeddingsirregularlyspaced} (2 heads, 5 layers). Finally, the event decoder computes the conditional intensity function for some future time by conditioning on the encoded history to produce probabilistic forecasts.\\
\newline
PatientTPP extends the AttNHP class from the EasyTPP library \cite{xue2024easytppopenbenchmarkingtemporal} with several domain-specific enhancements (Figure 2A). First, we incorporate time-invariant patient features (gender, race, and ``pseudo-age'', the patient's age as of 1 January 2007) by injecting these static covariates into the model at the first time step of each sequence, allowing them to condition the trajectory from the outset. Second, we model numeric features (e.g., BMI, HbA1c) separately using a dedicated TPP stream. These continuously-valued inputs, which tend to occur at a higher cadence than that of categorical clinical events, are quantized into five discrete buckets per feature and processed via a parallel TPP submodel. The submodel's outputs are concatenated with the main model's final representation. Third, we integrate pretrained clinical embeddings of medical concepts derived from external sources that capture semantic similarity among related diagnoses, lab tests, and medications \cite{johnson2024unified}. Embeddings are determined by a fixed mapping from our conditions of interest to the clinical embedding topics (Supp. Data 2). The vectors associated with any matches are averaged and concatenated to the outputs of the event encoder and attention layers, enriching the model's patient representation with domain knowledge. Together, these modifications enhance the existing neural TPP architecture to better represent patient-level medical data that requires personal context, irregularly contains important numeric data, and is laden with domain-specific knowledge.
\subsection*{Model training and evaluation}        
For the analysis under consideration, PatientTPP was trained and validated on 426,039 and 10,869 patient histories, respectively. Patients were required to have sequences of at least 32 indicative events (that is, exclusive of numeric features) and any events in excess of 64 were right-truncated. Histories with fewer than 64 events were right-padded with a padding token. The model is trained to maximize the log-likelihood of observed event types and times. This objective comprises (i) the log of the conditional intensity at each observed event and (ii) a survival term representing the probability of no event between observed events. Continuous-time modeling is supported by an adaptive thinning algorithm that samples candidate event times with an oversampling factor of 5 and draws up to 50 Monte Carlo samples per step when generating predictions. Model parameters used were: batch size 256, adam optimizer, and learning rate $10^{-3}$ for up to 100 epochs. Validation was performed every epoch, and the best-performing checkpoint was selected based on the validation set's negative log-likelihood. Root mean square error (RMSE) in the time domain and a sequence similarity metric (diff-ratio) in the event type domain were also tracked (Supp. Figure 1). For testing, we evaluated precision and recall for a subset of diagnoses: acute myocardial infarction (MI), subsequent myocardial infarction, complication of myocardial infarction, acute ischemic heart disease excluding MI, cardiac arrest, arrhythmias excluding atrial fibrillation, ischemic stroke, hemorrhagic stroke, atrial fibrillation, cardiomyopathy, cholelithiasis, cholecystitis, acute pancreatitis, esophageal cancer, stomach cancer, colorectal cancer, hepatocellular carcinoma, gallbladder cancer, pancreatic cancer, breast cancer , endometrial cancer, ovarian cancer, kidney cancer, meningioma, thyroid cancer, multiple myeloma, chronic kidney disease (CKD) stage 5 and ESRD, primary CKD stage 1-4, acute kidney failure, sleep apnea, chronic obstructive pulmonary disease (COPD), and patient death.

%figure 2
\begin{figure*}[tp]
\centering
\includegraphics[width=1.0\linewidth]{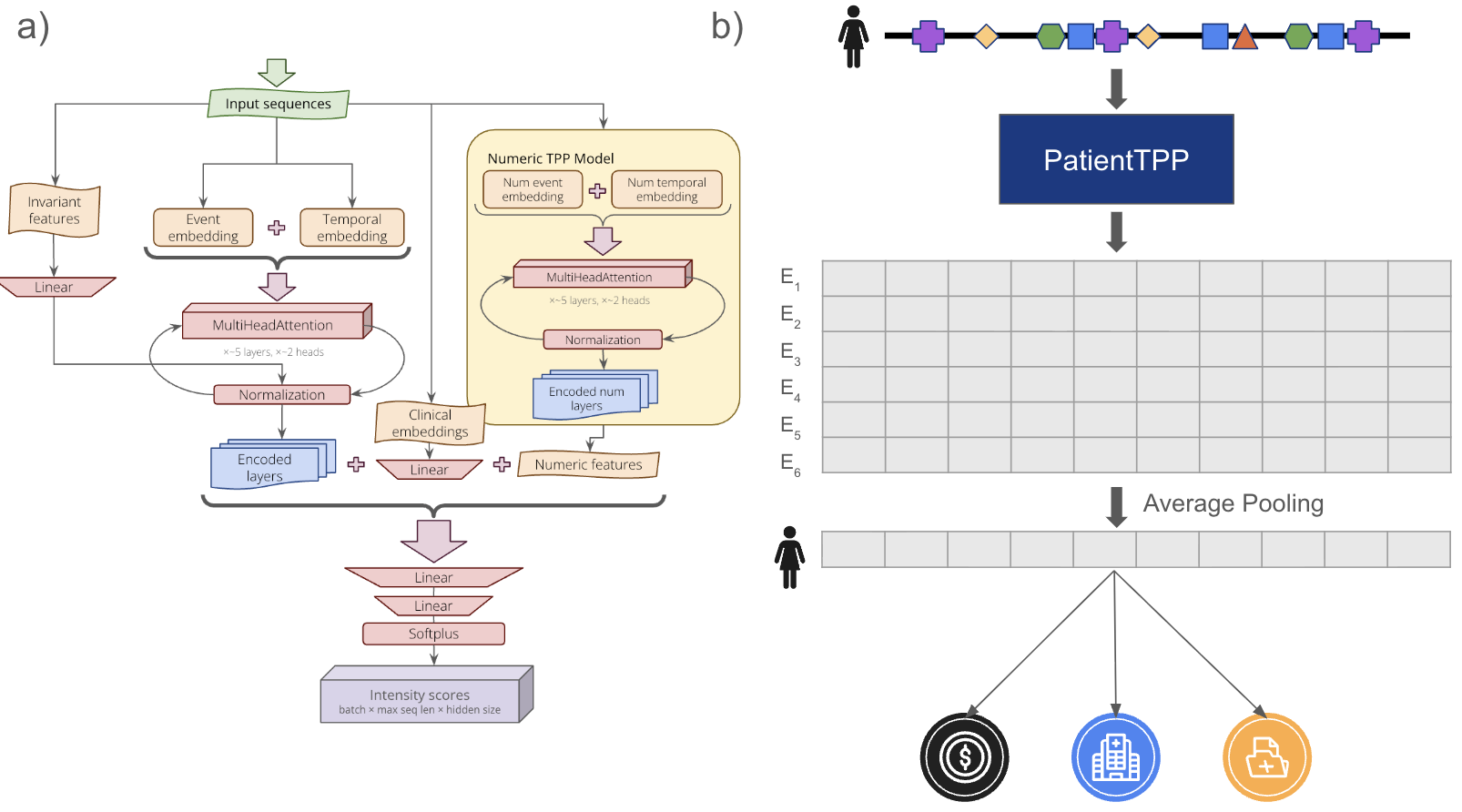}
\caption{\textbf{A.} PatientTPP extensions to the AttNHP encoder architecture. Input sequence events and timestamps are embedded separately, then concatenated and fed to a multi-head self-attentive transformer. Invariant features are added in the main context-building loop are are nonzero for the first item in the sequence. Numeric features pass through their own TPP submodel (yellow) and are concatenated with the indicative feature results and the clinical embeddings.  The unified stream is transformed by two fully connected layers and a Softplus activation, yielding intensity scores.  The output has dimensions $[batch size] {\times} [maximum sequence length] {\times} [hidden layer size]$. \textbf{B.} Patient characterization and TPP inference. Hidden layers of the trained TPP model encode the patient's context and functionally serve as an embedding space for EHR, simplifying cohort identification and risk assessment. We pool the contexts for six predicted conditions to arrive at a vector representation of a patient.}
\label{fig:model-topo}
\end{figure*}

\subsection*{Patient representations}       
Following training, PatientTPP produced, for a patient sequence history \(H\), a matrix of predicted event probabilities \(P \in \mathbb{R}^{|T| \times I}\), where each element \(P[e,i]\) represents the model’s estimate that event type \(e \in T\) occurs within time bin \(i \in \{1,\ldots,I\}\) under the causal mask. This probability grid provides a temporally resolved view of the model’s forecast for all modeled clinical event types. We then pooled probabilities across the forecast window for each event type to derive fixed-length patient representations from these outputs :
\[
\phi_e = \sum_{i=1}^{I} w_i \, P[e,i], \quad e \in T,
\]
where \(w_i\) are pooling weights over time bins. The resulting vector \(\Phi(H) = (\phi_e)_{e\in T}\) serves as a patient-level embedding summarizing the expected likelihood or frequency of each modeled clinical event (Figure 2B). Each coordinate \(\phi_e\) has direct semantic meaning: it represents the time-aggregated probability of observing event \(e\) over the prediction horizon. This construction yields embeddings that are both numerically stable and clinically interpretable, since each dimension aligns with a specific, named clinical concept. These patient embeddings form the basis for the supervised modeling task described bellow.
\subsection*{Overweight/obese cohort construction}
To construct a low-risk overweight/obese cohort for potential GLP-1 benefit, we first identified a GLP-1 treated cohort to match untreated patients. A semaglutide-treated cohort was defined using the following criteria. Inclusion criteria were: age $>=18$; BMI $>=25$; and semaglutide doses $>=1$  between April 2021 and March 2023. Exclusion criteria were: any history of diabetes or treatment with diabetes medication, end stage renal disease, or cancer (with the exception of non-melanoma skin cancer); any history of MI, stroke or peripheral arterial disease within one year prior to index; $<12$ months of claims data prior to or after semaglutide index date (except if the patient died); and history of weight loss drug use. Diagnoses and medications were profiled in patient histories by ICD-10-CM (Supp. Data 3) and NDC (Supp. Data 4) codes, respectively. The resulting cohort of 68,968 patients was then matched with patients who meet all inclusion and exclusion criteria with the exception of the semaglutide treatment inclusion criterion using propensity score matching (PSM). For matching, covariates included patient baseline demographics, diagnosis, and medication history. Nearest neighbor matching was used to match 1:1 and success was considered a standardized mean differences $SMD < 0.1$ for the matched cohorts \cite{KOMATSOULIS2025100587}. Six mutually exclusive, non-treated cohorts were generated by PSM. These patients were not used in PatientTPP training.\\
\newline
Patients in the cohorts were processed by PatientTPP to generate patient-level embeddings. Each patient was assigned an index date that represented the last date of observation for PatientTPP inference. Index dates were assigned based on the distribution of semaglutide start dates in the reference, treated cohort. PatientTPP predicted the first six future events for each patient. Intensities were normalized to $[0,1]$ over all events for all patients and each event vector was normalized to $[0,1]$. Patient embeddings were generated by averaging the six event vectors for each patient. Intensities for fixed events `gender', `race', and `pseudo-age' were removed from the final embeddings for a 1 x 62 dimension embedding per patient.
\subsection*{Obesity outcomes}
For all overweight/obese patients, patient histories were profiled for outcomes by the presence of curated ICD-10-CM codes (Supp. Data 3). Logistic regression models were trained per outcome on PatientTPP embeddings over five of the six cohorts and tested on the holdout cohort. If the number of negative patients for an outcome was greater than 5x the number of positive patients, the negatives were downsampled to 5x. Models were trained with scikitlearn LogisticRegression with max\_iter=1000 and class\_weight=`balanced' \cite{scikit-learn}. Performance was measured as the area under the receiver operating curve (AUROC) and reported as the average over the six holdout tests. To identify intensities that were predictive of outcomes, we trained the logistic regression models as above but with elastic net regularization over stratified five-fold splits of the training cohort. We again tested with each of the six cohorts held out, and report the average weight and the number of times the feature was significantly predictive. We utilized scikit-learn LogisticRegressionCV with penalty=`elasticnet', l1\_ratios=0.75, solver=`saga', class\_weight=`balanced', max\_iter=1000, scoring=`precision' \cite{scikit-learn}.
\subsection*{Cost analysis}
For the same patients, claims data in Optum\textsuperscript{\textregistered} Market Clarity was profiled for post-index standardized cost (all-cost) for medical services and prescriptions. Cardiovascular-associated cost (CV-cost) was calculated for claims with procedure or NDC codes related to CV disease (Supp. Data 5 and Supp. Data 6, respectively). PatientTPP representations were used as input to a linear regression model trained with scikit-learn LinearRegression with default parameters \cite{scikit-learn}. Models were trained with the same cohort train/test procedure as above and performance was reported as average $R^2$. We calculated cumulative gain curves to evaluate the ability of PatientTPP-derived rankings to capture disproportionate shares of cost, a metric commonly used in payer risk modeling. Cumulative gain curves were calculated by ranking patients highest to lowest and then calculating the percent of total cohort cost captured for each additional patient down the ranking. Statistical significance was assessed using a paired permutation test (10,000 iterations) that randomly swapped the two ranking scores within each patient with 50\% probability and recomputed the metric difference. The one-sided p-value was estimated as the proportion of permutations in which the permuted difference was greater than or equal to the observed difference.

\section*{Results}
\subsection*{Model performance}
PatientTPP was tested on sequences of 106,510 patients by backtesting the last six events in the sequence. We were interested in the ability to predict previously undiagnosed conditions with significant morbidity (Table 1). We excluded patients that had one or more of these diagnoses in both the future event predictions and their history for a refined cohort of 63,901 patients. For each event prediction for each patient, we considered an event prediction a true positive if the prediction appeared in the held out history at any position of the six, while we considered it a false positive if it was predicted but not in the actual sequence. We considered an event present in the actual sequence but not predicted to be a false negative. While the average recall across all diagnoses was 10\%, precision was $>50\%$ for those diagnoses where predictions were made (Table 1). For comparison, we utilized weighted guessing based on the incidence rates of the events in the test sequences. An example patient trajectory that includes PatientTPP-generated future events is depicted in Supp. Figure 2.
\begin{table*}[!htbp]
\centering
\resizebox{\textwidth}{!}{%
\begin{tabular}{l|r|rrrrr|rrrrr}
\hline
\multirow{2}{*}{\textbf{Condition}} &
\multicolumn{6}{c|}{\textbf{PatientTPP}} &
\multicolumn{5}{c}{\textbf{Weighted guessing}} \\
\cline{2-12}
& \textbf{Support} & \textbf{TP} & \textbf{FP} & \textbf{FN} & \textbf{Prec.} & \textbf{Recl.}
& \textbf{TP} & \textbf{FP} & \textbf{FN} & \textbf{Prec.} & \textbf{Recl.} \\
\hline
None & 53635 & 52510 & 8466 & 1125 & 0.86 & 0.98 & 43462 & 8352 & 10173 & 0.84 & 0.81 \\
Sleep Apnea & 3029 & 460 & 835 & 2569 & 0.36 & 0.15 & 147 & 2796 & 2882 & 0.05 & 0.05 \\
Arrhythmias excl. atrial fibrillation & 1827 & 45 & 49 & 1782 & 0.48 & 0.02 & 54 & 1754 & 1773 & 0.03 & 0.03 \\
Primary CKD stage 1--4 & 1410 & 260 & 289 & 1150 & 0.47 & 0.18 & 20 & 1369 & 1390 & 0.01 & 0.01 \\
COPD & 1342 & 225 & 176 & 1117 & 0.56 & 0.17 & 18 & 1230 & 1324 & 0.01 & 0.01 \\
Atrial fibrillation & 952 & 252 & 205 & 700 & 0.55 & 0.26 & 14 & 883 & 938 & 0.02 & 0.01 \\
Acute Kidney Failure & 682 & 0 & 0 & 682 & NaN & 0.00 & 8 & 674 & 674 & 0.01 & 0.01 \\
Cholelithiasis & 540 & 0 & 1 & 540 & 0.00 & 0.00 & 4 & 527 & 536 & 0.01 & 0.01 \\
Patient death & 530 & 0 & 1 & 530 & 0.00 & 0.00 & 9 & 458 & 521 & 0.02 & 0.02 \\
Ischemic stroke & 496 & 33 & 13 & 463 & 0.72 & 0.07 & 4 & 472 & 492 & 0.01 & 0.01 \\
Acute Myocardial Infarction & 419 & 0 & 0 & 419 & NaN & 0.00 & 6 & 391 & 413 & 0.02 & 0.01 \\
Cardiomyopathy & 397 & 59 & 66 & 338 & 0.47 & 0.15 & 1 & 356 & 396 & 0.00 & 0.00 \\
Cholecystitis & 158 & 0 & 0 & 158 & NaN & 0.00 & 0 & 153 & 158 & 0.00 & 0.00 \\
Cardiac Arrest & 149 & 0 & 0 & 149 & NaN & 0.00 & 0 & 140 & 149 & 0.00 & 0.00 \\
Breast cancer & 105 & 35 & 40 & 70 & 0.47 & 0.33 & 0 & 103 & 105 & 0.00 & 0.00 \\
Acute Pancreatitis & 99 & 0 & 0 & 99 & NaN & 0.00 & 0 & 85 & 99 & 0.00 & 0.00 \\
Acute Ischemic Heart Disease excl. MI & 98 & 0 & 0 & 98 & NaN & 0.00 & 0 & 101 & 98 & 0.00 & 0.00 \\
Hemorrhagic stroke & 72 & 0 & 0 & 72 & NaN & 0.00 & 0 & 59 & 72 & 0.00 & 0.00 \\
Colorectal cancer & 42 & 4 & 10 & 38 & 0.29 & 0.10 & 0 & 51 & 42 & 0.00 & 0.00 \\
Kidney cancer & 28 & 4 & 1 & 24 & 0.80 & 0.14 & 0 & 22 & 28 & 0.00 & 0.00 \\
CKD 5 and ESRD & 25 & 4 & 9 & 21 & 0.31 & 0.16 & 0 & 31 & 25 & 0.00 & 0.00 \\
Pancreatic cancer & 25 & 1 & 0 & 24 & 1.00 & 0.04 & 0 & 24 & 25 & 0.00 & 0.00 \\
Endometrial cancer & 24 & 1 & 0 & 23 & 1.00 & 0.04 & 0 & 28 & 24 & 0.00 & 0.00 \\
Thyroid cancer & 22 & 1 & 1 & 21 & 0.50 & 0.05 & 0 & 25 & 22 & 0.00 & 0.00 \\
Hepatocellular carcinoma & 19 & 0 & 0 & 19 & NaN & 0.00 & 0 & 18 & 19 & 0.00 & 0.00 \\
Multiple myeloma & 13 & 0 & 0 & 13 & NaN & 0.00 & 0 & 15 & 13 & 0.00 & 0.00 \\
Ovarian cancer & 9 & 0 & 0 & 9 & NaN & 0.00 & 0 & 7 & 9 & 0.00 & 0.00 \\
Complication of MI & 8 & 0 & 0 & 8 & NaN & 0.00 & 0 & 6 & 8 & 0.00 & 0.00 \\
Subsequent MI & 8 & 0 & 0 & 8 & NaN & 0.00 & 0 & 7 & 8 & 0.00 & 0.00 \\
Stomach cancer & 8 & 0 & 0 & 8 & NaN & 0.00 & 0 & 8 & 8 & 0.00 & 0.00 \\
Esophageal cancer & 6 & 3 & 6 & 3 & 0.33 & 0.50 & 0 & 6 & 6 & 0.00 & 0.00 \\
Gallbladder cancer & n$<$5 & 0 & 0 & n$<$5 & NaN & 0.00 & 0 & 1 & n$<$5 & 0.00 & 0.00 \\
Meningioma & n$<$5 & 0 & 0 & n$<$5 & NaN & 0.00 & 0 & 2 & n$<$5 & 0.00 & 0.00 \\
\hline
\end{tabular}}
\caption{Performance comparison of PatientTPP vs. weighted guessing for new diagnosis prediction. Patients evaluated only have diagnoses present in the final six events of available histories. ``None'' represents an event not in the set of diagnoses of interest.}
\label{tab:tpp-classification-metrics}
\end{table*}
\subsection*{PatientTPP transfer learning}
While PatientTPP is an encoder-decoder model, we evaluated whether future event predictions could be used in transfer learning as a patient-level representation for future prediction tasks. To do this, we generated six future events for a patient and averaged the event vectors for a one-dimensional vector representation of a patient at a given time (Figure 2B).
\subsection*{Obesity-associated health outcomes}
To evaluate whether PatientTPP representations are useful for modeling obesity-associated health risk in otherwise low-risk individuals, we needed to identify patients who: 1) were overweight or obese; 2) had no history of diabetes, cardiovascular disease, end stage renal disease, or cancer; and 3) had no history of GLP-1 or weight loss medication use. In Optum® Market Clarity, 4,643,605 patients met these inclusion and exclusion criteria. To identify likely GLP-1 candidates, we selected patients by matching to a cohort that met the same low-risk definition but were treated with semaglutide. We generated six of these cohorts (Supp. Table 1). On average, PatientTPP was able to embed 49.5\% of each cohort for a total of 194,080 patients used in transfer learning experiments. Patients were observed for an average of 590 days after embedding (Supp. Figure 3) with events in that time period considered future events.\\
\newline
For a number of important obesity-associated health outcomes, including cardiovascular, metabolic, and non-metabolic diagnoses, we trained logistic regression models on PatientTPP representations. The incidence of most outcomes prior to index date across cohorts was low with the exception of hypertension and hyperlipidemia, which were almost 50\%, and the number of patients with each outcome increased in the post index period (Supp. Table 2). Each regression was trained on five of the six cohorts and tested on the holdout cohort. The model was able to beat a random classifier for all outcomes with AUROC ranging from 0.53-0.79 (Figure 3). Post-index MI and stroke events are new for these patients as CVD history was an exclusion criteria for the cohort, therefore, classification model performances represent the ability to predict new MIs and strokes. Models could discriminate for outcomes that were not included as events in PatientTPP training, including pulmonary embolism, deep vein thrombosis, metabolic dysfunction-associated steatotic liver disease, osteoarthritis, gastroesophageal reflux disease, gout, and mood disorder.\\
\newline
To gain insight into the latent features that drove individual outcome prediction, we utilized logistic regression coefficients learned for each outcome. We examined the most influential PatientTPP representation features associated with two outcomes, MI and ischemic stroke (Figure 4). For MI, predictors with strong positive coefficients included elevated event intensities for eGFR testing, acute kidney failure, complication of MI, and acute MI. Negative coefficients were linked to intensities for total cholesterol testing, HbA1c testing, subsequent MI, and weight observation. Similarly, ischemic stroke was most strongly predicted by event intensities for chronic kidney disease, acute MI, and cholelithiasis, while negative predictors were strong intensities for subsequent MI and testing for triglyceride, HbA1c, eGFR, and LDL levels.
% figure 3
\begin{figure*}[!bhtp]
\centering
\includegraphics[width=1.0\textwidth]{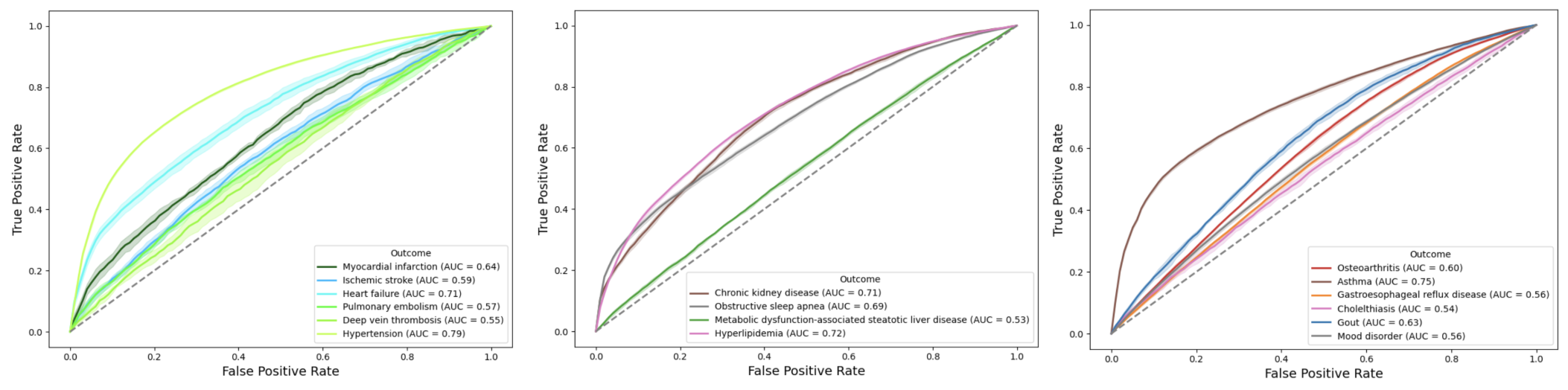}
\caption{Outcome prediction from PatientTPP representations. For each outcome, a binary logistic regression was trained on PatientTPP representations for the presence or absence of the outcome in the period after embedding. Outcomes are organized into cardiovascular (left), metabolic (center), or other (right). Average area under the curve is reported for 6-fold hold-out test.}
\label{fig:obesity-outcomes}
\end{figure*}
% figure 4
\begin{figure*}[!bhtp]
\centering
\includegraphics[width=1.0\textwidth]{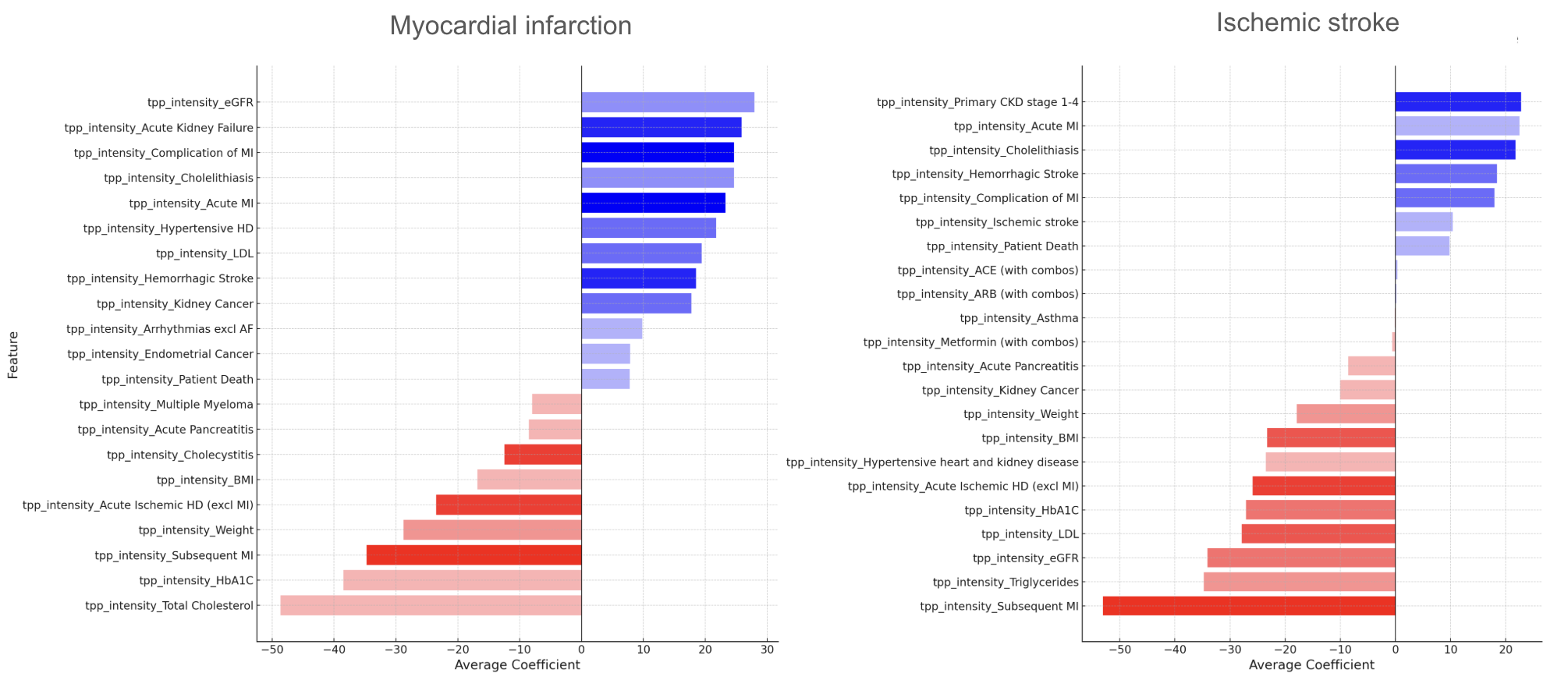}
\caption{Feature importance for outcome prediction. Each bar represents a regression coefficient associated with a PatientTPP embedding dimension corresponding to a specific clinical event type (y-axis). These dimensions reflect the learned intensity of events used during TPP model training. Only features with nonzero coefficients for the given outcome are shown. Bar color indicates coefficient sign (blue = positive association, red = negative), and shading reflects consistency: darker bars appeared in more of the testing folds (max = 6).}
\label{fig:feature-importance}
\end{figure*}
\subsection*{Healthcare risk}
Ultimately, obesity-associated risk is a summation over multiple, disparate potential outcomes. To assess whether PatientTPP representations capture healthcare burden, we looked at future healthcare utilization and cost. As above, we trained logistic regression models on five of six cohorts and tested on the holdout. Average AUROC for any hospitalization was 0.56 and any emergency room visit was 0.57 with slight improvement identifying patients with multiple visits (Supp. Figure 4). For cost, we investigated all-cost and CV-cost (Supp. Figure 5). The model could not learn a relationship for all-cost (average $R^2=0.01$) but was able to learn a relationship with CV-cost (average $R^2=0.22$).\\
\newline
Given that BMI is commonly used as a decision cutoff for GLP-1 treatment for weight loss, we used BMI stratification to compare against PatientTPP-derived CV-cost (Figure 5A). We compared both rankings using cumulative gain curves, a standard evaluation in payer risk modeling. Predicted CV-cost achieved a 17\% higher area under the cumulative cost capture curve (Figure 5B), representing a 36\% relative improvement over BMI ($p<10^{-5}$). The top 25\% of patients ranked by PatientTPP accounted for 39.3\% of total cost, compared to only 22.7\% when ranking by BMI. We selected a cohort size equal to the fraction of patients with $BMI > 35.0$ representing a potential GLP-1 treatment cohort ($n=82,963$, $47.6\%$). The difference in average patient cost was \$975 higher when ranked by PatientTPP-derived CV-cost compared to BMI (Supp. Figure 6), or \$80.9 million across the potential cohort, within up to 4 years of observation.\\
% figure 5
\begin{figure*}[!bhtp]
\centering
\includegraphics[width=0.8\textwidth]{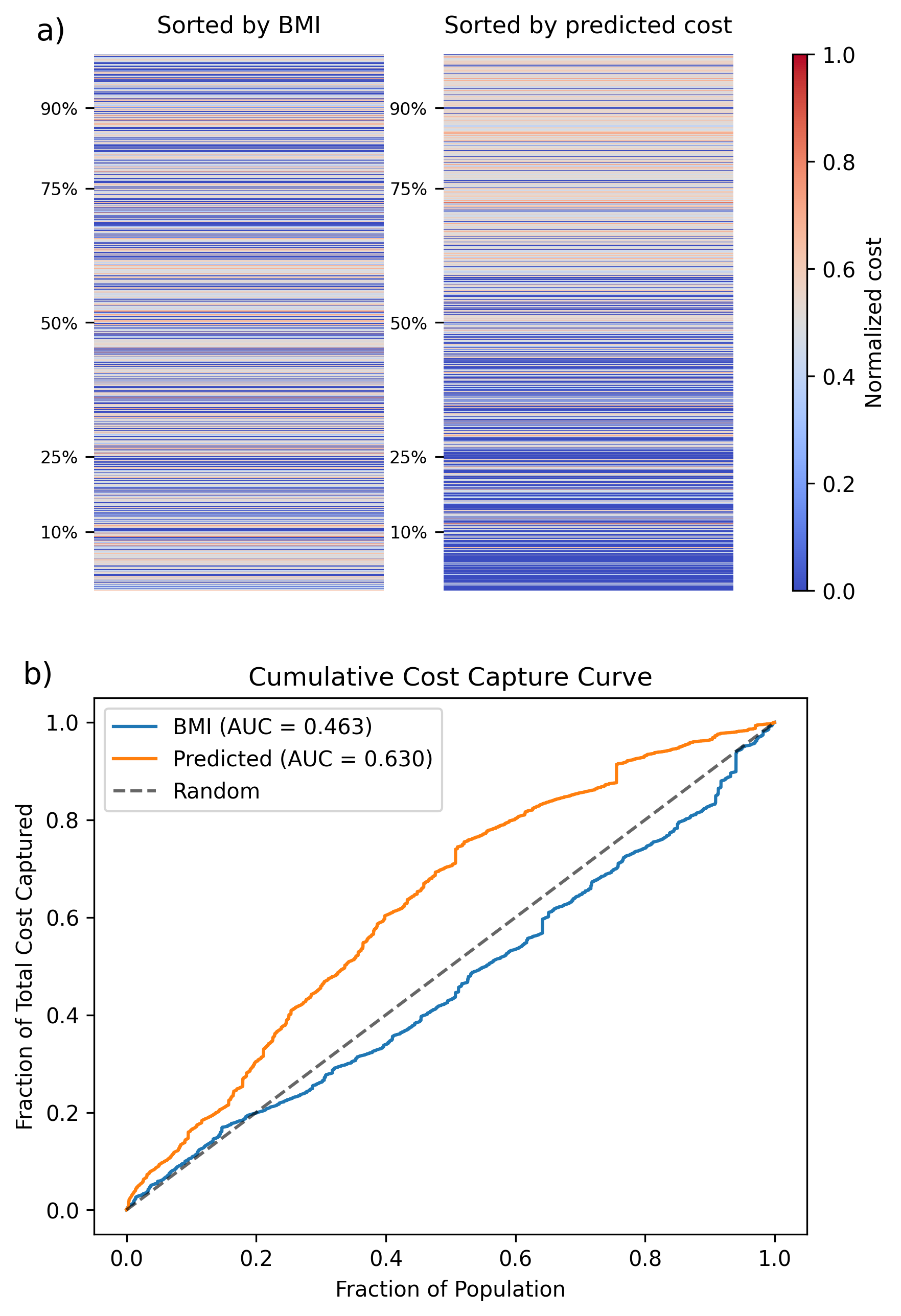}
\caption{Comparison of BMI and PatientTPP-predicted CV costs for patient stratification. \textbf{A.} Future CV-associated costs for all low-risk overweight/obese cohorts were ranked by BMI and CV-cost predicted from PatientTPP representations. \textbf{B.} Cumulative cost capture curve for the same patients when ranked by BMI and predicted CV-cost.}
\label{fig:cum-cost-capture}
\end{figure*}

\section*{Discussion}
% figure 6
\begin{figure*}[tbh]
\centering
\includegraphics[width=1.0\textwidth]{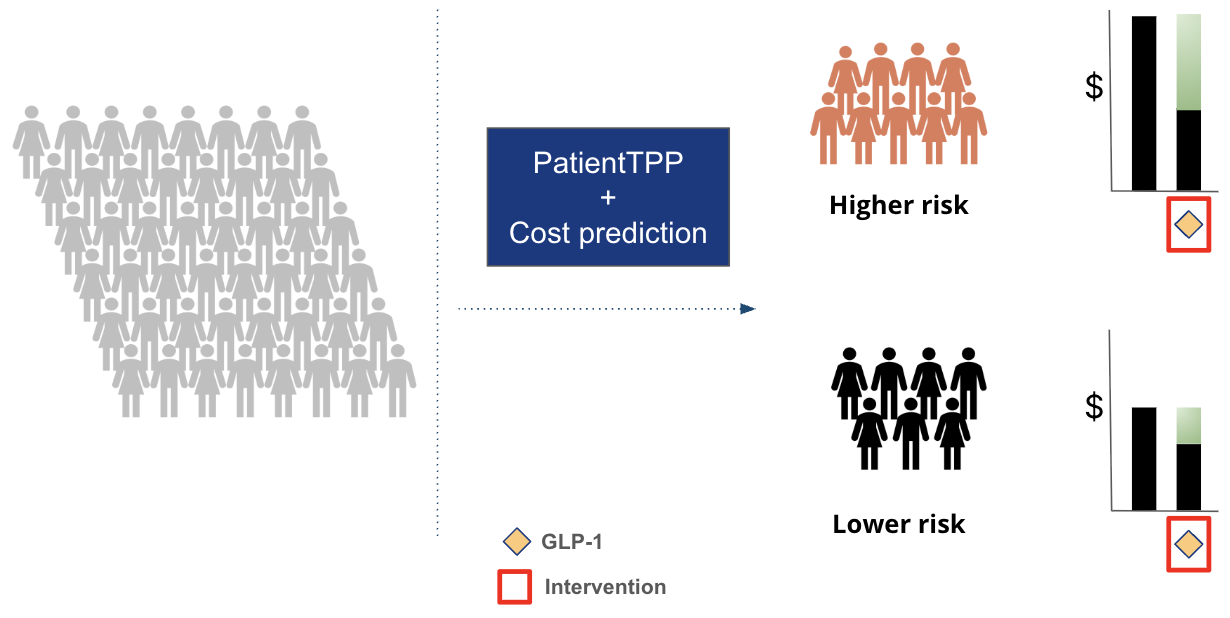}
\caption{Population risk stratification with PatientTPP. Using TPP model representations for low-risk overweight and obese patients we identified a higher burden cohort for potential treatment with GLP-1 therapy. A similar approach can be used to stratify for other high cost or high risk interventions using PatientTPP.}
\label{fig:patient-stratification}
\end{figure*}
In healthcare, the ability to risk stratify is imperative to both providers and payers. This is especially true when considering high stakes decisions, whether they be invasive or high cost. When considering longitudinal risk associated with chronic conditions, of which overweight/obesity has the highest prevalence in the U.S., the ability to discern differences between similar looking patients is near impossible with statistical models that rely on limited, engineered features. The recent success of novel GLP-1s in decreasing risk across a growing number of adverse outcomes of having a high BMI has made the risk modeling problem more acute. The potentially benefiting population is huge, the cost of the drugs are high, but the reduction in adverse outcomes is only realized by a relative few. As has been acknowledged, this imbalance has led to decreased coverage of these medications for weight loss \cite{Klein2024}. One solution to this societal challenge is better risk modeling for patients with obesity (Figure 6).\\
\newline
At the level of an individual patient, AI can make a difference in preventive care when it can discern differences in patients where traditional clinical evaluation does not. Our work demonstrates that patient representations learned via neural temporal point processes capture strong predictive signals across both cardiovascular and non-cardiovascular outcomes associated with obesity. Importantly, these include endpoints that were explicitly modeled during TPP training as well as novel outcomes that were not, indicating that the learned patient representations are robust and transferable. The framework also shows predictive signal for outcomes relevant to health economics and outcomes research, including future healthcare cost stratification, underscoring its utility for both clinical and operational applications.\\
\newline
While TPP models offer a powerful framework for modeling longitudinal health data, their application to healthcare modeling has been limited due to imperfect alignment with real-world patient data. PatientTPP provides solutions for some such challenges. First, TPPs require precise timestamps to capture event dynamics; however, clinical timestamps are often coarse or noisy due to batching or retrospective entry, which limits temporal resolution and introduces uncertainty. To mitigate this, we adopted quarterly time binning, which trades off precision for robustness. Second, their inability to natively incorporate time-invariant or continuous covariates, such as demographic characteristics or time-varying lab values, which are often strong predictors of clinical outcomes. To address this, we integrate auxiliary encoders for static and numeric features and augment the TPP framework to condition on these representations. Additionally, TPPs are computationally intensive, particularly during inference and training. By using PatientTPP to generate a patient representation, we constrain inference to a small number of TPP passes per patient. Finally, evaluating TPPs can be less intuitive than traditional classification models because of the time component to event generation. To assess performance, we utilized a transfer learning framework that enabled evaluation of PatientTPP on downstream, clinically meaningful classification and risk prediction tasks that we assessed with traditional evaluation metrics.\\
\newline
There is a problem in machine learning, especially as it applies to clinical modeling, that learned representations do not correspond elements of the representation to interpretable features. Neuronal TPP modeling offers some mitigation of this challenge by corresponding each element of the embedding to the event type it weights for future event prediction. We note that the dimensions of the PatientTPP representation that are predictive of future MI and ischemic stroke had a strong clinical relationship with these outcomes. Crucially, the patients used in these two classification tasks did not have these events in their histories, meaning the signal captured in the PatientTPP for these events, that is then useful in predicting their future outcome, is derived from other aspects of their medical histories. While event weights provide signal as to what elements of a history are predictive of a future event, further work is necessary to understand the relationship between histories, embedding event weights, and future event arrivals.\\
\newline
We acknowledge the following important limitations of this study. First, PatientTPP was trained, tested, and evaluated in a single, U.S.-based dataset, which may limit generalizability. Second, we restricted the set of event types to a curated group we deemed important for future risk in overweight or obese patients; however, there are likely other event types that could improve model performance if included. Additionally, our transfer learning framework utilized learned event embeddings but did not leverage the time-to-event forecasting capabilities of TPPs. Similarly, our future event prediction task asked whether a patient experienced an outcome within their follow-up period but did not model time-to-event explicitly. Future work is necessary to develop an evaluation scheme for time forecasting of clinical events with low incidence. Finally, while we demonstrate that PatientTPP-derived representations are predictive of a range of outcomes and costs, prospective evaluation is needed to determine the real-world utility and clinical impact of trajectory-based risk stratification for guiding the use of high-cost interventions such as GLP-1 therapy.

\section*{Institutional review board (IRB) review}
This retrospective observational study used fully de-identified data from the Optum® Market Clarity database. No direct contact with subjects occurred, and the dataset contains no individually identifiable information (e.g., names, addresses, or medical record numbers). Therefore, the research does not meet the definition of human subjects research under 45 CFR 46.102 and is exempt from IRB review per 45 CFR 46.104 (d)(4). Accordingly, institutional review board approval was not required.

\section*{Acknowledgments}
This study was sponsored by Zephyr AI.

\section*{Author contributions}
Z.N.F., D.T., and J.S. conceived of the project and designed the study. X.Y.Z. and H.P. contributed to study design. R.K., J.W., N.L., and N.T. organized and processed the data. Z.N.F. and D.T. conducted experiments, analyzed results, and produced figures. Z.N.F. drafted the manuscript. Z.N.F., D.T., H.P., and J.S. revised the manuscript. All authors provided critical feedback for the final manuscript.

\section*{Competing interests}
Zachary N. Flamholz reports a relationship with Zephyr AI that includes: equity or stocks. Dillon Tracy reports a relationship with Zephyr AI that includes: equity or stocks. Ripple Khera reports a relationship with Zephyr AI that includes: equity or stocks. Jordan Wolinsky reports a relationship with Zephyr AI that includes: equity or stocks. Nicholas Lee reports a relationship with Zephyr AI that includes: equity or stocks. Nathaniel Tann reports a relationship with Zephyr AI that includes: equity or stocks. Xiao Yin Zhu reports a relationship with Zephyr AI that includes: equity or stocks. Harry Phillips reports a relationship with Zephyr AI that includes: equity or stocks. Jeffrey Sherman reports a relationship with Zephyr AI that includes: equity or stocks. If there are other authors, they declare that they have no known competing financial interests or personal relationships that could have appeared to influence the work reported in this paper.

\section*{Data Availability}
Patient TPP training data is available with an Optum® Market Clarity data license. PatientTPP model weights are available by request. Supplementary Figures 1–6 and Supplementary Tables 1–2 are provided in the Supplementary Information PDF. Supplementary Data 1–6 are provided as a separate Excel file.

\section*{Code Availability}
We make available the PatientTPP code and a synthetic dataset of patient clinical trajectories for training PatientTPP models in a github repository: \\
https://github.com/zephyr-ai-public/patient-tpp/.

%----------------------------------------------------------------------------------------
%	REFERENCE LIST
%----------------------------------------------------------------------------------------
\phantomsection
\bibliographystyle{unsrt2authabbrvpp}
\bibliography{references}
\clearpage

% Extended Data Figure 1 - FIXME
\setcounter{figure}{0}
\begin{figure}[p]
\centering
\includegraphics[width=1.0\textwidth]{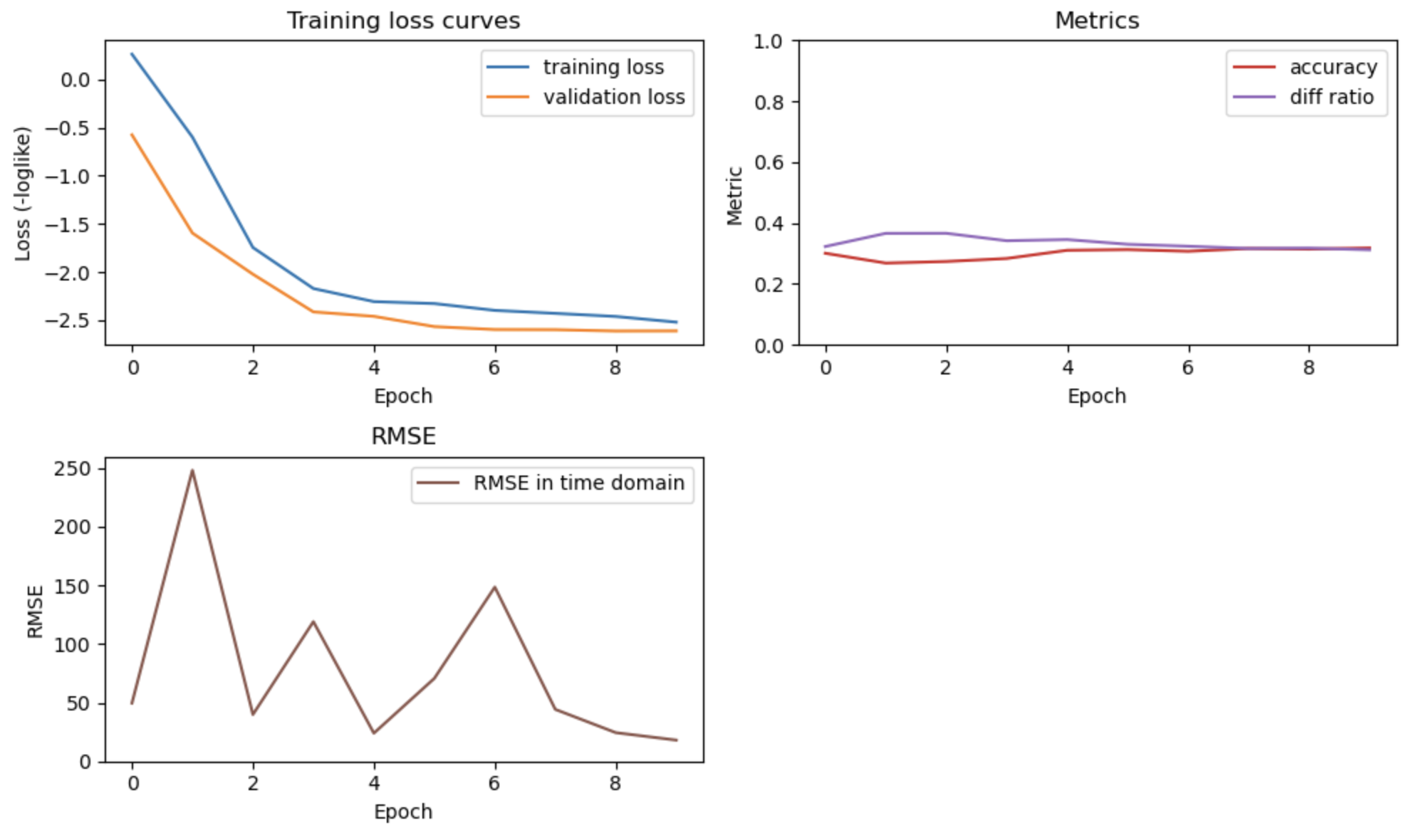}
\caption*{\textbf{Supplemental Figure 1.} Top left- Loss curves for training and validation sets. Top right- Accuracy and diff ratio metrics on the validation set (indicative events only). Accuracy is the fraction of events having correctly predicted type in the correct position, for 6 positions; diff ratio is a $[0, 1]$ similarity measure proportional to the number of matched events in two sequences and inversely proportional to the sequence length. These metrics depend on sampling the predicted joint distribution and have limited utility. Bottom left- root mean square error of predicted interevent times. All plots refer to semaglutide model training.}
\label{fig:supp-tpp-training}
\end{figure}
\clearpage

% Extended Data Figure 2
\begin{figure}[p]
\centering
\includegraphics[width=1.0\textwidth]{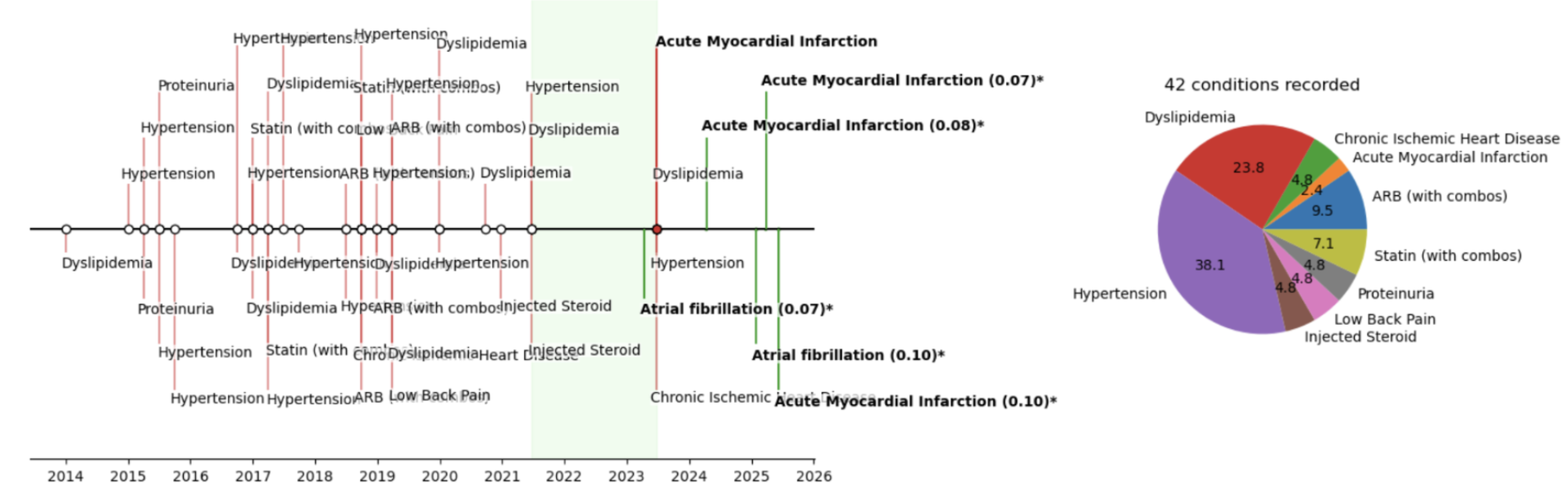}
\caption*{\textbf{Supplemental Figure 2.} Visualizing a patient timeline. Left- patient with 42 events of interest over roughly 8 years. The green-shaded backtesting region is defined by the last six events in the sequence. Events occurring prior to the backtesting window (red stems) are used to set the model context, and predictions are then made incrementally (green stems, bold text). Note that the model has not seen this sample in training, and neither myocardial infarction nor atrial fibrillation is in the patient history. Right- breakdown of patient conditions, exclusive of predictions.}
\label{fig:supp-patient-timeline}
\end{figure}
\clearpage

% Extended Data Figure 3
\begin{figure}[p]
\centering
\includegraphics[width=1.0\textwidth]{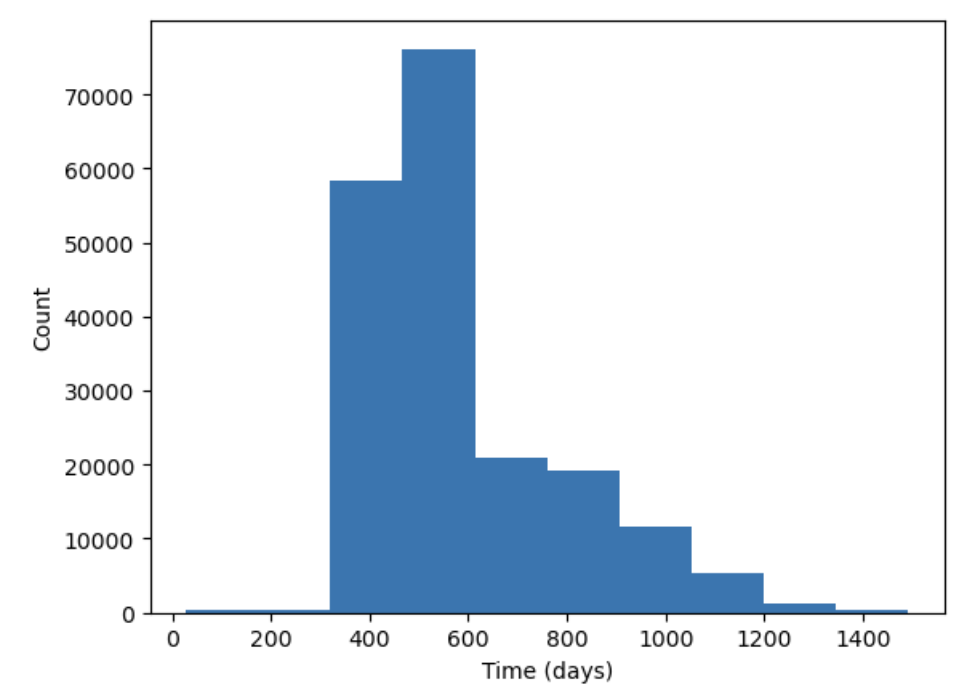}
\caption*{\textbf{Supplemental Figure 3.} Distribution of patient observation time after index date. The follow up window is used for future outcome prediction.}
\label{fig:supp-observation-time}
\end{figure}
\clearpage

% Extended Data Figure 4
\begin{figure}[p]
\centering
\includegraphics[width=1.0\textwidth]{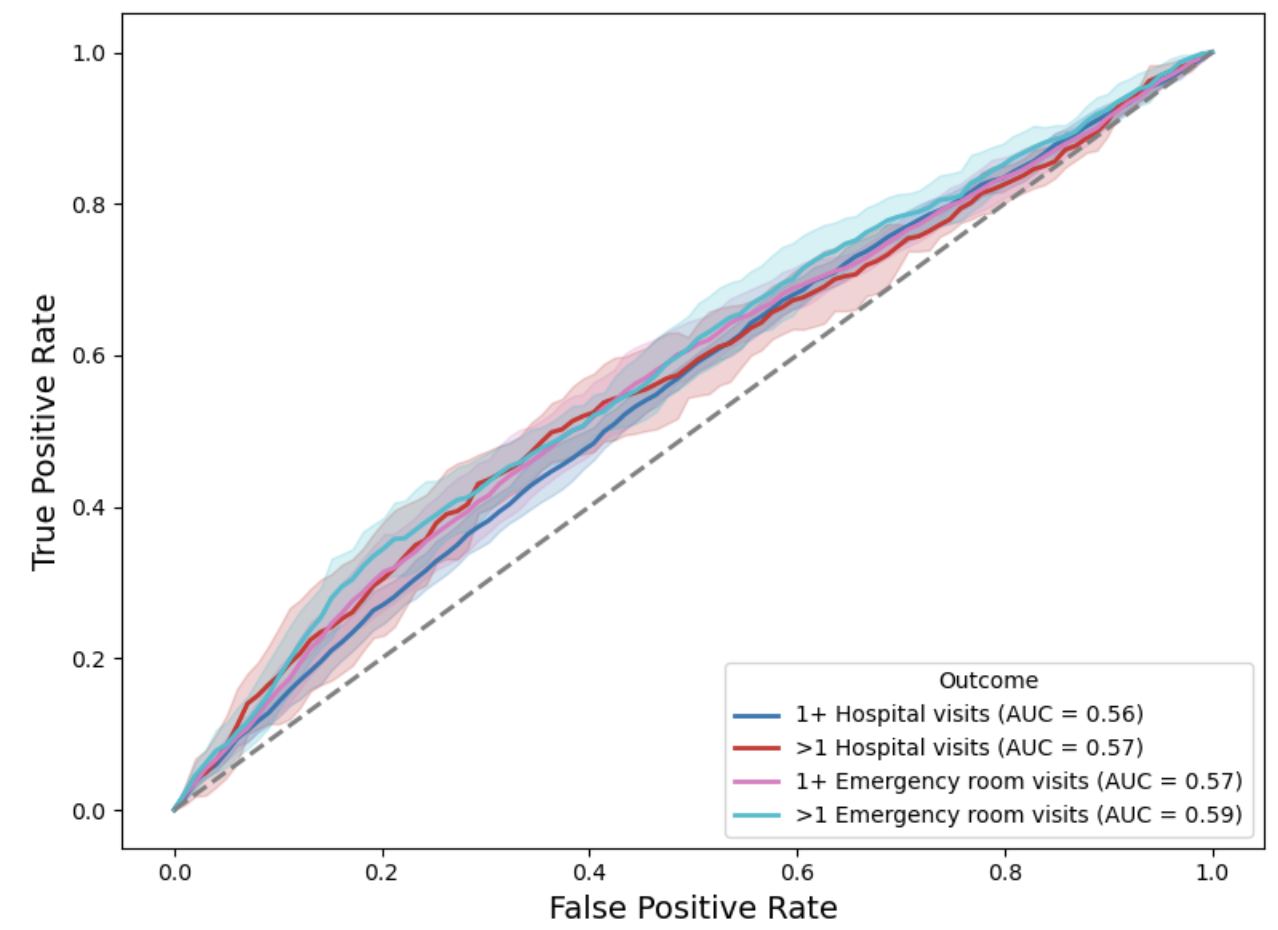}
\caption*{\textbf{Supplemental Figure 4.} ROC curves for prediction of emergency room visits and hospitalizations in the post-index period.}
\label{fig:supp-heor}
\end{figure}
\clearpage

% Extended Data Figure 5 - FIXME
\begin{figure}[p]
\centering
\includegraphics[width=1.0\textwidth]{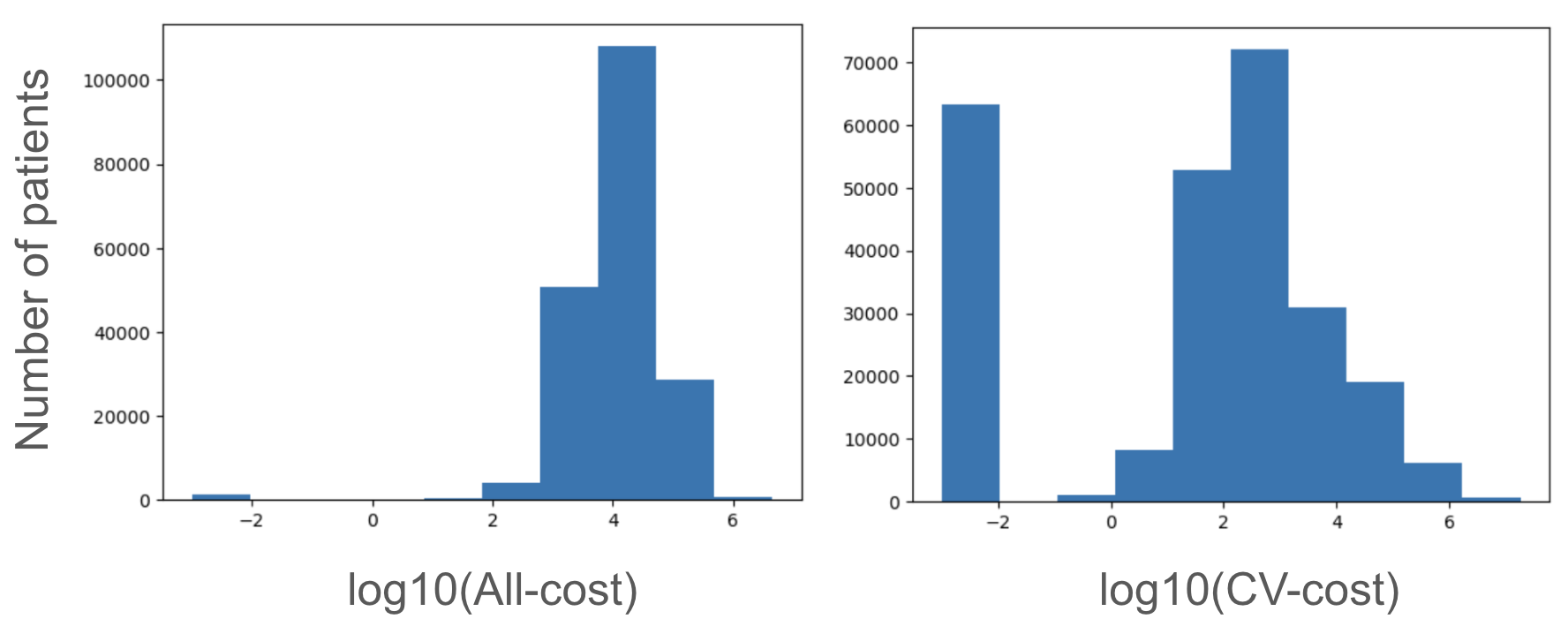}
\caption*{\textbf{Supplemental Figure 5.} Distribution of costs incurred by patients in the semaglutide cohort, over all conditions (left) and over cardiovascular conditions (right).}
\label{fig:supp-fig-cost-distributions}
\end{figure}
\clearpage

% Extended Data Figure 6
\begin{figure}[p]
\centering
\includegraphics[width=1.0\textwidth]{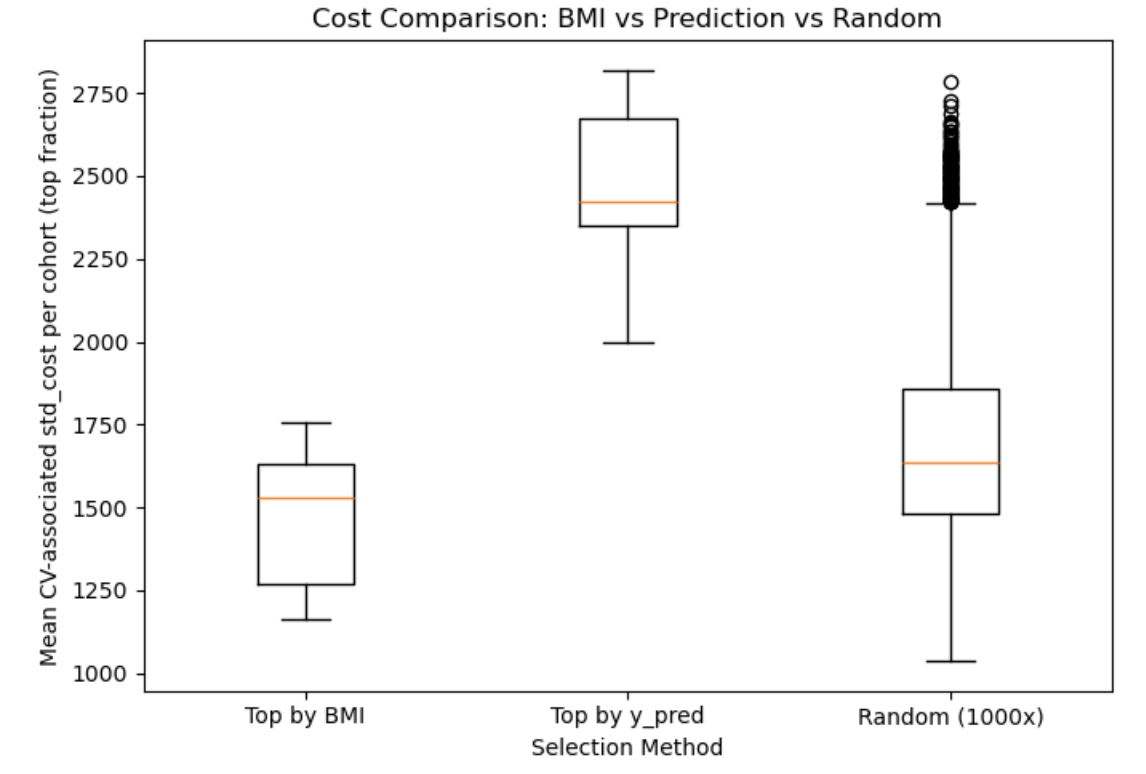}
\caption*{\textbf{Supplemental Figure 6.} Comparing real post-index CV-cost for a theoretical GLP-1 treatment cohort of size ($n=82,963$) based on the fraction of low-risk cohort patients with BMI $>35$. Distribution represents top \textit{n} when ranked by each method. For random, cohorts of size \textit{n} were evaluated for mean cost per patient, and this procedure was done 100 times.}
\label{fig:supp-fig-real-cost-bmi-frac}
\end{figure}
\clearpage

%% TABLES- https://docs.google.com/spreadsheets/d/14MCM_zOCpccMLeeFV-4C3VyP6D2XRHAA7lYG2bXgQSk/edit?usp=sharing

% Supplemental Table 1
% tab:supp-tpp-event-types = TPP event types

% Supplemental Table 2
% tab:supp-clinical-knowledge-embeddings = Clinical embedding aggregations

% Supplemental Table 3
% tab:supp-icd-10s = Obesity outcome ICD codes

% Supplemental Table 4
% tab:supp-ndcs = Obesity cohort medication NDCs

% Supplemental Table 5
% tab:supp-cv-procedure-codes = CV-associated drug codes

% Supplemental Table 6
% tab:supp-cv-ndc-codes = CV-asscoated procedure codes

% Supplemental Table 7
\begin{table}[ht]
\centering
\begin{tabular}{lcc}
\hline
\textbf{Cohort} & \textbf{Number of patients} & \textbf{Patients with representations} \\
\hline
1 & 66{,}607 & 33{,}570 \\
2 & 66{,}247 & 33{,}473 \\
3 & 65{,}852 & 33{,}029 \\
4 & 65{,}377 & 32{,}558 \\
5 & 64{,}506 & 29{,}681 \\
6 & 63{,}443 & 31{,}767 \\
\hline
\end{tabular}
\caption*{\textbf{Supplemental Table 1.} Patient counts and availability of PatientTPP representations by cohort.}
\label{tab:supp_tab_1}
\end{table}

% Supplemental Table 8
\begin{table}[ht]
\centering
\begin{tabular}{p{0.40\linewidth}ccc}
\hline
\textbf{Outcome} &
\textbf{Pre-index (n, \%)} &
\textbf{Post-index (n, \%)} &
\textbf{Change (n)} \\
\hline
Asthma & 29{,}462 (15.18) & 35{,}572 (18.33) & 6{,}110 \\
Cholelithiasis & 3{,}801 (1.96) & 5{,}888 (3.03) & 2{,}087 \\
Chronic kidney disease & 7{,}358 (3.79) & 10{,}986 (5.66) & 3{,}628 \\
Deep vein thrombosis & 1{,}650 (0.85) & 2{,}491 (1.28) & 841 \\
Gastroesophageal reflux disease & 42{,}822 (22.07) & 54{,}972 (28.33) & 12{,}150 \\
Gout & 4{,}051 (2.09) & 5{,}408 (2.79) & 1{,}357 \\
Heart failure & 3{,}615 (1.86) & 6{,}057 (3.12) & 2{,}442 \\
Hyperlipidemia & 82{,}533 (42.52) & 99{,}551 (51.28) & 17{,}018 \\
Hypertension & 94{,}108 (48.49) & 106{,}629 (54.96) & 12{,}521 \\
Metabolic dysfunction-associated steatotic liver disease &
9{,}541 (4.92) & 14{,}761 (7.61) & 5{,}220 \\
Myocardial infarction & 0 (0.00) & 1{,}445 (0.74) & 1{,}445 \\
Mood disorder & 56{,}853 (29.30) & 65{,}137 (33.55) & 8{,}284 \\
Obstructive sleep apnea & 31{,}217 (16.08) & 39{,}978 (20.60) & 8{,}761 \\
Osteoarthritis & 36{,}164 (18.64) & 50{,}249 (25.90) & 14{,}085 \\
Pulmonary embolism & 1{,}384 (0.71) & 1{,}988 (1.02) & 604 \\
Stroke & 0 (0.00) & 1{,}872 (0.96) & 1{,}872 \\
\hline
\end{tabular}
\caption*{\textbf{Supplemental Table 2.} Pre- and post-index prevalence of clinical outcomes and absolute change in counts.}
\label{tab:pre_post_outcomes}
\end{table}

\newpage
\noindent\textbf{Supplemental Data 1.} Clinical event types used for modeling PatientTPP.\\
\textbf{Supplemental Data 2.} Pre-trained embedding aggregations for PatientTPP event types.\\
\textbf{Supplemental Data 3.} International Classification of Diseases, 10th Revision, Clinical Modification (ICD-10-CM) codes used to identify obesity-associated outcomes.\\
\textbf{Supplemental Data 4.} National Drug Code (NDC) codes used to identify drugs of interest.\\
\textbf{Supplemental Data 5.} Cardiovascular disease-associated drug classes and drug names.\\
\textbf{Supplemental Data 6.} Cardiovascular disease-associated procedure codes from the International Classification of Diseases, 10th Revision, Procedure Coding System (ICD-10-PCS).

\clearpage

%----------------------------------------------------------------------------------------

\end{document}